\documentclass[twocolumn]{article}
\usepackage[T1]{fontenc}
\usepackage[margin=1in]{geometry}
\usepackage{authblk}
\usepackage{graphicx}
\usepackage{comment}
\usepackage{xcolor}
\usepackage{amsmath}
\usepackage{adjustbox}
\usepackage[utf8]{inputenc}

\title{The Hidden Costs of Translation Accuracy: Distillation, Quantization, and Environmental Impact}

\author[1]{Dhaathri Vijay}
\author[2]{Anandaswarup Vadapalli}

\affil[1]{University of California, Santa Cruz, \texttt{dhaathrivijay@gmail.com}}
\affil[2]{Research Spark Hub Inc, \texttt{anandaswarup.vadapalli@gmail.com}}

\begin{document}
\maketitle

\begin{abstract}
The rapid expansion of large language models (LLMs) has heightened concerns about their computational and environmental costs. This study investigates the trade-offs between translation quality and efficiency by comparing full-scale, distilled, and quantized models using machine translation as a case study. We evaluated performance on the Flores+ benchmark and through human judgments of conversational translations in French, Hindi, and Kannada.
Our analysis revealed that the full 3.3B FP32 model, while achieving the highest BLEU scores, incurred the largest environmental footprint ($\approx$ 0.007–0.008 kg $\text{CO}_{2}$ per run). The distilled 600M FP32 model reduced inference time by 71–78\% and carbon emissions by 63–65\% compared with the full model, with only minimal reductions in BLEU scores. Human evaluations further showed that even aggressive quantization (INT4) preserved high levels of accuracy and fluency, with differences between models generally minor.
These findings demonstrate that model compression strategies can substantially reduce computational demands and environmental impact while maintaining competitive translation quality, though trade-offs are more pronounced in low-resource settings. We argue for evaluation frameworks that integrate efficiency and sustainability alongside accuracy as central dimensions of progress in NLP.
\end{abstract}

\section{Introduction}
Natural Language Processing (NLP), a subfield of artificial intelligence at the intersection of linguistics and machine learning, has advanced rapidly with the rise of large language models (LLMs). Transformer based LLMs, trained on internet scale corpora, have shifted the field from narrow, task specific systems to general-purpose models capable of reasoning, code generation, and open ended language use~\cite{devlin2019bert, chowdhery2022palm, touvron2023llama}.

These advances, however, come with substantial computational and environmental costs. Training a single large scale model has been estimated to emit several hundred tons of $\text{CO}_{2}$, comparable to the lifetime emissions of multiple cars~\cite{zewe2025generativeai}. Such figures underscore the need for approaches that maintain strong performance while mitigating environmental impact.

Two widely studied strategies for improving efficiency are distillation and quantization. In distillation, a smaller student model learns to approximate a larger teacher model, while in quantization, the numerical precision of model parameters is reduced. Together, these techniques enable more efficient deployment of LLMs, particularly in resource constrained environments.

In this work, we examine how distillation and quantization affect the balance between translation quality and efficiency in both high- and low-resource languages. Although high-resource languages such as French have been extensively evaluated, low-resource languages such as Kannada remain underexplored, despite their greater vulnerability to performance degradation. By systematically comparing full, distilled, and quantized models, we highlight the trade-offs between translation quality, inference speed, and carbon emissions, and situate our findings within broader discussions of sustainable and responsible NLP.

\section{Related Work}
A large body of research has focused on improving the efficiency of large language models while preserving performance.

Early work on distillation demonstrated its effectiveness in producing compact yet accurate models. For example, DistilBERT~\cite{sanh2019distilbert} applied knowledge distillation during pretraining rather than task-specific fine-tuning. By using a composite loss function that combined language modeling, distillation, and cosine distance objectives, the authors reduced BERT’s parameters by 40\% while retaining 97\% of its accuracy. More recently, Hsieh et al.~\cite{hsieh2023distilling} proposed step-by-step distillation, in which student models were trained with rationale-based auxiliary supervision. Their 770M parameter T5 model outperformed a 540B parameter PaLM model on few-shot tasks while using only 80\% of the data. Other lightweight approaches, such as MiniALBERT~\cite{nouriborji2022minialbert}, combined distillation with parameter sharing across layers to achieve further efficiency gains. Collectively, these studies show that distillation can substantially compress models while preserving much of their performance.

Quantization has also emerged as a powerful tool for reducing memory footprint and computation. Zhang et al.~\cite{zhang2022post} introduced a generalized post-training quantization method with theoretical guarantees, showing that networks could be quantized to as few as 4 bits per weight with minimal loss in accuracy. Wei et al.~\cite{wei2023outlier} addressed the performance degradation caused by outliers in transformer channels by proposing Outlier Suppression+ (OS+), which achieved near-floating-point performance on models such as BERT, OPT, BLOOM, and LLaMA.

Despite extensive research, most prior work has studied distillation and quantization independently and has primarily focused on high-resource languages. Our study builds on this by jointly examining the two approaches in the context of machine translation, with special attention to low-resource languages where performance trade-offs are often most severe.

\section{Methodology}

We evaluated translation quality and efficiency using a combination of automatic benchmarks and human judgments. This dual framework allowed us to assess both computational performance and translation quality.

\subsection{Models Considered}
\label{subsec:models}
We investigated the performance of the No Language Left Behind (NLLB-200)~\cite{nllb-24} family of models across different parameter counts and precision levels. The following models were included in our study:
\begin{enumerate}
\item \texttt{nllb\_3.3b\_fp32}: NLLB 3.3B parameter model with full precision (FP32).
\item \texttt{nllb\_600M\_fp32}: 600M parameter full precision (FP32) model distilled from  \texttt{nllb\_3.3b\_fp32}
\item \texttt{nllb\_600M\_fp16 / nllb\_600M\_int8 / nllb\_600M\_int4}: FP16, INT8 and INT4 quantized variants of the 600M parameter model.
\end{enumerate}

All experiments were implemented in PyTorch, making use of pretrained NLLB-200 models from the HuggingFace Transformers library. The experiments were executed on a single 40GB A100 GPU hosted on Google Colab. To ensure comparability across runs, GPU memory was explicitly cleared between experiments using Python garbage collection and PyTorch CUDA utilities. The results were systematically recorded in a structured Pandas DataFrame for subsequent analysis.

These models allowed us to compare the effects of distillation (3.3B → 600M) and quantization (FP32 → FP16/INT8/INT4) on translation performance and efficiency.

\subsection{Benchmark Evaluation on Flores+}
We first evaluated translation quality on the Flores+~\cite{nllb-24} devtest split, focusing on French, Hindi, and Kannada. Flores+ provides parallel corpora across more than 200 languages, making it one of the most comprehensive multilingual benchmarks. French serves as a representative high-resource language, while Hindi and Kannada serve as representative mid- and low-resource languages respectively.

Translation quality was measured using BLEU scores against reference translations. Efficiency was measured by average latency per sentence and carbon emissions estimated using CodeCarbon~\cite{benoit_courty_2024_11171501}.

\subsection{Human Evaluation}

To complement the objective scores, we constructed a dataset of 100 conversational English sentences. Each sentence was translated into French, Hindi, and Kannada by all five model variants. Language-proficient raters then evaluated translations along two criteria: accuracy (fidelity to source meaning) and fluency (grammaticality and naturalness). Both criteria were scored on a 1–5 scale, as summarized in Table~\ref{tab:rubric}.

\begin{table}[!htb]
\centering
\caption{Human evaluation rubric for translation accuracy and fluency (1–5 scale).}
\label{tab:rubric}
\vspace{0.5cm}
\begin{adjustbox}{width=0.48\textwidth}
\begin{tabular}{|c|l|l|}\hline
Score & Accuracy & Fluency \\ \hline
1 & Completely incorrect, & Incoherent, \\
 & mistranslation or & ungrammatical, \\
 & unrelated output & unnatural phrasing \\ \hline
2 & Major errors, large & Very awkward,  \\
 & portions of meaning & frequent grammar \\
 & lost & errors \\ \hline
3 & Partially accurate, & Understandable \\
& idea conveyed with & but awkward \\
& notable errors & \\ \hline
4 & Mostly accurate, &  Mostly fluent, \\
 & minor errors & minor phrasing  \\
 & & issues \\ \hline
5 & Fully accurate & Fully fluent \\ \hline
\end{tabular}
\end{adjustbox}
\end{table}

This rubric ensured consistent and transparent human evaluation, providing a subjective complement to objective metrics. 

\section{Results}
This section reports findings from both the benchmark-based evaluation and the human assessment of translations. Results are presented in two parts: (i) automatic evaluation on the Flores+ dataset and (ii) human evaluation of a custom dataset.

\subsection{Benchmark Evaluation (Flores+)}
Table~\ref{tab:benchmark-results} reports BLEU scores, inference times, and carbon emissions for the \texttt{nllb\_3.3b\_fp32} and the \texttt{nllb\_600M\_fp32} models across English-French, English-Hindi, and English-Kannada. The \texttt{nllb\_3.3b\_fp32} model achieved the highest BLEU scores in all language pairs. The performance gap between the full and distilled models was largest for English-Kannada (3.5 BLEU points), smaller for English-French (1.9 points), and smallest for English-Hindi (1.3 points). This pattern suggests that distillation disproportionately affected the low-resource language (Kannada), while high-resource French remained relatively robust.

Despite these reductions in BLEU scores, the distilled model achieved substantial efficiency gains. Inference was 71–78\% faster across all three language pairs, with average sentence latency dropping from 0.187s to 0.041s for English-French and from 0.146s to 0.042s for English-Kannada. Carbon emissions per evaluation run decreased by 63–65\%, from 0.0075-0.0079 kg $\text{CO}_2$ to 0.0025-0.0029 kg $\text{CO}_2$. These results demonstrate that distillation significantly reduced computational and environmental costs, albeit with modest quality degradation.

\begin{table*}[!ht]
\centering
\caption{Benchmark evaluation on Flores+ showing BLEU scores, inference time (in secs/sentence), carbon emissions per inference run (in kg $\text{CO}_2$) and relative percentage reductions for \texttt{nllb\_3.3b\_fp32} (Full) and \texttt{nllb\_600M\_fp32} (Distilled) models.}
\label{tab:benchmark-results}
\vspace{0.5cm}
\begin{tabular}{|l|c|c|c|c|c|c|c|c|}
\hline
& \multicolumn{2}{|c|}{BLEU Score} & \multicolumn{3}{|c|}{Inference Time} & \multicolumn{3}{|c|}{Carbon Emissions} \\ \hline
Language Pair & Full & Distilled & Full & Distilled & Reduction(\%) & Full & Distilled & Reduction(\%) \\ \hline
English-French & 43.99 & 42.10 & 0.187 & 0.041 & 78.1\% & 0.0075 & 0.0026 & 65.3\% \\ 
English-Kannada & 15.31 & 11.80 & 0.146 & 0.042 & 71.2\%& 0.0079 & 0.0029 & 63.3\% \\ 
English-Hindi & 28.67 & 27.42 & 0.134 & 0.037 & 72.4\% & 0.0069 & 0.0025 & 63.8\% \\ \hline
\end{tabular}
\end{table*}

\subsection{Human Evaluation of Custom Dataset}
Human ratings of accuracy and fluency (Table~\ref{tab:human-eval}) showed that all models performed at a generally high level, with mean scores above 4.2. 

The \texttt{nllb\_3.3b\_fp32} model achieved the highest ratings for French (accuracy 4.8, fluency 4.75), aligning with its BLEU advantage in this high-resource setting. However, differences among the distilled and quantized variants were small, typically within 0.2 points.

Interestingly, the \texttt{nllb\_600M\_int4} model obtained the highest accuracy in Hindi (4.47) while maintaining competitive fluency across languages. This result suggests that quantization did not uniformly degrade performance and, in some cases, may have improved alignment with human judgments.

Overall, human evaluations confirmed the benchmark findings: distilled and quantized models retained strong translation quality, with only minor losses relative to the full model.

\begin{table*}[!ht]
\centering
\caption{Mean accuracy and fluency ratings (1-5 scale) across languages and model variants.}
\label{tab:human-eval}
\vspace{0.5cm}
\begin{tabular}{|l|c|c|c|c|c|c|}
\hline
& \multicolumn{3}{|c|}{Accuracy} & \multicolumn{3}{|c|}{Fluency} \\
\hline
Model & Hindi & Kannada & French & Hindi & Kannada & French \\
\hline
\texttt{nllb\_3.3b\_fp32} & 4.41 & 4.43 & 4.80 & 4.29 & 4.33 & 4.75 \\
\texttt{nllb\_600M\_fp32} & 4.36 & 4.41 & 4.74 & 4.23 & 4.41 & 4.67 \\
\texttt{nllb\_600M\_fp16} & 4.37 & 4.45 & 4.70 & 4.24 & 4.45 & 4.65 \\
\texttt{nllb\_600M\_int8} & 4.34 & 4.45 & 4.73 & 4.19 & 4.43 & 4.69 \\
\texttt{nllb\_600M\_int4} & 4.47 & 4.43 & 4.58 & 4.34 & 4.42 & 4.55 \\
\hline
\end{tabular}
\end{table*}

\section{Conclusion}
This study examined the trade-offs among translation quality, computational efficiency, and environmental impact for large-scale machine translation models. We compared the NLLB-3.3B full-precision model, its 600M distilled variant, and quantized versions across three languages of differing resource availability.

The distilled models reduced inference latency by 71–78\% and carbon emissions by 63–65\% relative to the full model, while incurring only modest losses in BLEU scores. Performance degradation was most pronounced for low-resource Kannada, whereas high-resource French remained largely stable.

Human evaluations confirmed that even aggressive quantization preserved high levels of accuracy and fluency, with differences among models generally small.

These results demonstrate that distillation and quantization can substantially cut computational demands and environmental costs without severely compromising translation quality. However, the sharper trade-offs observed in low-resource languages highlight the need for careful evaluation before deployment in linguistically diverse settings.

Our analysis focused on inference-time efficiency; the training phase, which contributes significantly to the total carbon footprint of large models, remains an important direction for future research. Expanding to a broader set of languages and tasks will further illuminate multilingual trade-offs and advance the agenda of responsible and environmentally conscious NLP.


\bibliographystyle{ieeetr}
\bibliography{references}

\begin{thebibliography}{10}

\bibitem{devlin2019bert}
J.~Devlin, M.-W. Chang, K.~Lee, and K.~Toutanova, ``Bert: Pre-training of deep
  bidirectional transformers for language understanding,'' {\em Proceedings of
  NAACL-HLT}, 2019.

\bibitem{chowdhery2022palm}
A.~Chowdhery, S.~Narang, J.~Devlin, M.~Bosma, G.~Mishra, A.~Roberts, P.~Barham,
  H.~W. Chung, Q.~Le, {\em et~al.}, ``Palm: Scaling language modeling with
  pathways,'' {\em arXiv preprint arXiv:2204.02311}, 2022.

\bibitem{touvron2023llama}
H.~Touvron, T.~Lavril, G.~Izacard, X.~Martinet, M.-A. Lachaux, T.~Lacroix,
  B.~Rozi{\`e}re, N.~Goyal, E.~Hambro, F.~Azhar, {\em et~al.}, ``Llama: Open
  and efficient foundation language models,'' {\em arXiv preprint
  arXiv:2302.13971}, 2023.

\bibitem{zewe2025generativeai}
A.~Zewe, ``Explained: Generative ai’s environmental impact,'' {\em MIT News},
  January 2025.

\bibitem{sanh2019distilbert}
V.~Sanh, L.~Debut, J.~Chaumond, and T.~Wolf, ``Distilbert, a distilled version
  of bert: smaller, faster, cheaper and lighter,'' {\em arXiv preprint
  arXiv:1910.01108}, pp.~1798--1828, 2019.

\bibitem{hsieh2023distilling}
C.-Y. Hsieh, C.-L. Li, C.-K. Yeh, H.~Nakhost, Y.~Fujii, A.~Ratner, R.~Krishna,
  C.-Y. Lee, and T.~Pfister, ``Distilling step-by-step! outperforming larger
  language models with less training data and smaller model sizes,'' {\em arXiv
  preprint arXiv:2305.02301}, 2023.

\bibitem{nouriborji2022minialbert}
M.~Nouriborji, O.~Rohanian, S.~Kouchaki, and D.~A. Clifton, ``Minialbert: Model
  distillation via parameter-efficient recursive transformers,'' {\em arXiv
  preprint arXiv:2210.06425}, 2022.

\bibitem{zhang2022post}
J.~Zhang, Y.~Zhou, and R.~Saab, ``Post-training quantization for neural
  networks with provable guarantees,'' {\em arXiv preprint arXiv:2201.11113},
  2022.

\bibitem{wei2023outlier}
X.~Wei, Y.~Zhang, Y.~Li, X.~Zhang, R.~Gong, J.~Guo, and X.~Liu, ``Outlier
  suppression+: Accurate quantization of large language models by equivalent
  and optimal shifting and scaling,'' {\em arXiv preprint arXiv:2304.09145},
  2023.

\bibitem{nllb-24}
{NLLB Team}, M.~R. Costa-juss{\`a}, J.~Cross, O.~{\c{C}}elebi, M.~Elbayad,
  K.~Heafield, K.~Heffernan, E.~Kalbassi, J.~Lam, D.~Licht, J.~Maillard,
  A.~Sun, S.~Wang, G.~Wenzek, A.~Youngblood, B.~Akula, L.~Barrault, G.~M.
  Gonzalez, P.~Hansanti, J.~Hoffman, S.~Jarrett, K.~R. Sadagopan, D.~Rowe,
  S.~Spruit, C.~Tran, P.~Andrews, N.~F. Ayan, S.~Bhosale, S.~Edunov, A.~Fan,
  C.~Gao, V.~Goswami, F.~Guzm{\'a}n, P.~Koehn, A.~Mourachko, C.~Ropers,
  S.~Saleem, H.~Schwenk, and J.~Wang, ``Scaling neural machine translation to
  200 languages,'' {\em Nature}, vol.~630, no.~8018, pp.~841--846, 2024.

\bibitem{benoit_courty_2024_11171501}
B.~Courty, V.~Schmidt, S.~Luccioni, Goyal-Kamal, MarionCoutarel, B.~Feld,
  J.~Lecourt, LiamConnell, A.~Saboni, Inimaz, supatomic, M.~Léval, L.~Blanche,
  A.~Cruveiller, ouminasara, F.~Zhao, A.~Joshi, A.~Bogroff, H.~de~Lavoreille,
  N.~Laskaris, E.~Abati, D.~Blank, Z.~Wang, A.~Catovic, M.~Alencon,
  M.~Stęchły, C.~Bauer, L.~O.~N. de~Araújo, JPW, and MinervaBooks,
  ``mlco2/codecarbon: v2.4.1,'' May 2024.

\end{thebibliography}
\end{document}